\documentclass[lettersize,journal]{IEEEtran}
\usepackage{amsmath,amsfonts}
\usepackage{array}
\usepackage[caption=false,font=normalsize,labelfont=sf,textfont=sf]{subfig}
\usepackage{textcomp}
\usepackage{stfloats}
\usepackage{url}
\usepackage{verbatim}
\usepackage{graphicx}
\usepackage{cite}
\usepackage{amssymb}
\usepackage[ruled,noline,linesnumbered,nofillcomment]{algorithm2e}
\usepackage{multirow}
\usepackage{makecell}
\usepackage{adjustbox} 
\usepackage{bbding}
\usepackage{pifont}
\usepackage{booktabs}
\usepackage{threeparttable}
\usepackage{graphicx}
\usepackage{url}
\begin{document}

\title{UNIVERS: Learning UNIversal Voxel-wise Embedding via Randomized Synthesis for generalizable one-shot landmark localization}

\author{Derong Yu, Qingsong Zhao, Yanbing Yue, Yikai Li, Zhi Zhang, and Guoyan Zheng
\thanks{This work was partially supported by National Natural Science Foundation of China via project 62471293. (Corresponding authors: Zhi Zhang and Guoyan Zheng.)}
\thanks{Derong Yu and Guoyan Zheng (email: guoyan.zheng@sjtu.edu.cn) are with the Institute of Medical Robotics, School of Biomedical Engineering, Shanghai Jiao Tong University, Shanghai 200240, China.}
\thanks{Qingsong Zhao, Yanbing Yue, Yikai Li, and Zhi Zhang (email: zhi.zhang@shgh.cn) are with the Department of Cardiology, Shanghai General Hospital (originally named Shanghai First People’s Hospital), Shanghai Jiao Tong University School of Medicine, Shanghai 200080, China.}
}

%

\maketitle

\begin{abstract}
One-shot landmark localization in volume scans with varying fields of view (FOVs) is challenging. While existing self-supervised learning methods have shown promising performance, they rely on contrastive learning built upon augmented views of the same image, which inherently limits their generalization in cross-modality scenarios. This motivates research on cross-modality embedding learning using paired multi-modality images. However, existing methods suffer from not only limited training data, but also the requirement of multi-stage iterative training. To address these limitations, we propose UNIVERS, an end-to-end trainable framework that learns UNIversal Voxel-wise Embedding via Randomized Synthesis. Our framework leverages diverse pseudo-data synthesized from organ shape templates, with randomization applied to semantic layout, visual appearance, FOV, and spatial pose. Coupled with cross-domain discriminative voxel-wise contrastive learning, this domain randomization strategy forces the network to disregard domain-specific biases and learn features that are robust against variations in imaging contrast, resolution, and spatial context. Consequently, the trained model enables one-shot landmark localization in multi-modality volume scans with arbitrary FOVs. In this study, the efficacy of the proposed method is demonstrated by applying it to the task of localizing the adrenal glands, which are paired, small, and irregular endocrine organs. In particular, comprehensive experiments are conducted on three challenging adrenal gland datasets. The experimental results demonstrate that UNIVERS consistently outperforms the state-of-the-art (SOTA) one-shot landmark localization approaches, with particularly prominent advantages in cross-modality scenarios. The source code is publicly available at \url{https://github.com/Kendra9999/UNIVERS}.
\end{abstract}


\section{Introduction}
\label{sec:introduction}
The adrenal glands are paired, small, irregular endocrine organs located superior to both kidneys \cite{lyraki2021adrenal}. By secreting essential hormones, they play vital roles in regulating stress response, metabolism, electrolyte homeostasis, and sympathetic nervous system \cite{fayemiwo2025novel,luo2021optimized}. Adrenal disorders, such as adrenal hyperplasia and adrenal tumor, are frequently encountered in routine radiological imaging \cite{mete2022overview}. Such abnormalities can disrupt hormonal balance, resulting in hypertension and other hormone-related symptoms \cite{fassnacht2016management}. Computer-aided interpretation of adrenal imaging offers a noninvasive and effective approach, playing an increasingly critical role in the clinical management of adrenal-related diseases. Due to the extremely small size of the adrenal glands, precise localization of their centers in volumetric medical images becomes an essential preprocessing step for many downstream computer-aided tasks, such as adrenal segmentation \cite{luo2021optimized} and lesion classification \cite{yang2025localization}. Existing landmark localization methods mostly rely on supervised learning \cite{payer2019integrating,ertl2026nnlandmark}, which requires large amounts of manually annotated training data. Unfortunately, annotating medical images is laborious, costly, and requires clinical expertise.

To reduce the annotation burden, recent research has shifted towards one-shot landmark localization, which requires only a single annotated support image to identify corresponding landmarks in query images \cite{lei2021contrastive,yan2022sam,lei2025medlsam,bai2025uae}. An emerging paradigm to find such landmark correspondences is achieved by learning discriminative voxel-wise anatomical embeddings. Typically, two augmented, overlapping views are extracted from the same image and fed into a Siamese network to generate voxel-wise embeddings. Then, self-supervised contrastive learning is applied, which encourages corresponding voxels across the two views to have similar embeddings while pushing non-corresponding voxels apart in the embedding space. Given the anatomical structural consistency, the learned embeddings map corresponding landmarks to similar feature representations. Consequently, annotations from a single support image can be transferred to query images via feature similarity matching. Building on this concept, several strategies have been proposed to enhance landmark localization performance, including negative sampling \cite{yan2022sam}, coarse-to-fine architectures \cite{lei2021contrastive,yan2022sam,lei2025medlsam}, and semantic embedding learning \cite{bai2025uae}. More recently, Universal Anatomical Embedding (UAE-M) \cite{bai2025uae} has extended this line of work to cross-modality landmark matching between Computed Tomography (CT) and Magnetic Resonance Imaging (MRI). This method learns voxel correspondences across unaligned multi-modality images via an iterative training process.

Despite significant progress, several limitations persist. First, most existing methods \cite{lei2021contrastive,yan2022sam,lei2025medlsam} build upon the heuristic that corresponding voxels across augmented views of the same image should yield similar embeddings, while their training is restricted to single-modality data. As a result, these methods struggle to match corresponding anatomical structures across multi-modality images. Second, while UAE-M \cite{bai2025uae} addresses cross-modality landmark matching, it relies on paired multi-modality images of the same subject. Such image pairs are usually unaligned across modalities, necessitating multi-stage iterative training to establish reliable voxel correspondences. More importantly, existing methods \cite{lei2021contrastive,yan2022sam,lei2025medlsam,bai2025uae} rely heavily on real clinical images, yet it is challenging, if not impossible, to collect large-scale multi-modality datasets. Accordingly, networks are exposed to a limited spectrum of domain variations in anatomical structure, imaging contrast, spatial resolution, field of view (FOV), and patient pose. Such insufficient data diversity results in suboptimal feature representations that lack robustness to large domain shifts in real clinical scenarios.  

To address these limitations, we propose UNIVERS, an end-to-end trainable framework that learns UNIversal Voxel-wise Embedding via Randomized Synthesis for generalizable one-shot landmark localization. Instead of relying on real training images, our learning process utilizes synthetic data generated from organ shape templates, where randomization is introduced in semantic layout, visual appearance, FOV, and spatial pose. This strategy exposes the network to a wide range of variability during contrastive learning, forcing it to learn features that are robust to variations in imaging contrast, resolution, and spatial context. Discriminative voxel-wise embedding learning is then achieved via a Cross-domain Discriminative Voxel-wise Contrastive (CDVC) loss, which encourages the voxel-wise discriminability of domain-invariant features amidst contextual variations. During inference, UNIVERS relies exclusively on anatomical structural consistency to distinguish voxels, generating domain-invariant voxel-wise embeddings for real clinical images from multiple domains without retraining or finetuning. UNIVERS achieves superior performance compared with existing state-of-the-art (SOTA) methods under three one-shot evaluation settings: intra-dataset, cross-dataset intra-modality, cross-dataset cross-modality. Our main contributions are summarized as follows:
\begin{itemize}
\item We propose UNIVERS, an end-to-end trainable framework for learning UNIversal Voxel-wise Embedding via Randomized Synthesis. To the best of our knowledge, this is the first work to learn generalizable, discriminative voxel-wise embeddings exclusively from pseudo-data synthesized via domain randomization.
\item We design a novel Cross-domain Discriminative Voxel-wise Contrastive (CDVC) loss, which operates on multi-domain pseudo-patch quadruplets with multiple positives to drive the network toward learning domain-invariant and discriminative embeddings.
\item We develop a coarse-to-fine one-shot adrenal gland localization framework built upon a single frozen UNIVERS network, which enables landmark localization in multi-modality volume scans with arbitrary FOVs.
\item We conduct comprehensive experiments under three one-shot settings across three distinct datasets. Experimental results demonstrate that our proposed UNIVERS consistently outperforms the SOTA approaches.
\end{itemize}

\section{Related work}

\subsection{One-shot landmark localization}
\label{subsec_oneshot}
To mitigate the annotation burden, one-shot landmark localization has garnered increasing research interest. Traditional solutions rely on image registration to establish voxel correspondences, typically employing a two-stage alignment pipeline: a global affine or rigid transformation \cite{feldmar1996rigid} followed by a deformable registration \cite{balakrishnan2019voxelmorph,chen2022transmorph,siebert2024convexadam}. By deriving a deformation field that warps the query image into the support image space, registration methods can propagate support landmarks to the query image. However, such methods often incur high computational costs and struggles with large inter-subject variations, including anatomical deformations and FOV shifts.

Alternatively, a growing body of work has explored embedding-learning-based paradigms on 3D CT images \cite{lei2021contrastive,yan2022sam,lei2025medlsam,bai2025uae}. These methods seek to learn discriminative voxel-wise embeddings, so that corresponding landmarks between the support and query images can be matched based on feature similarity. A representative framework is Self-supervised Anatomical eMbedding (SAM) \cite{yan2022sam}, which employs a coarse-to-fine architecture and designs a hard and diverse sampling strategy for effective negative selection. However, SAM primarily captures contextual appearance information and lacks explicit semantic understanding, which hinders its ability to differentiate organs with similar appearances. This limitation motivated UAE-S \cite{bai2025uae}, which extends SAM with an additional semantic head trained via a prototypical supervised contrastive loss. Another line of work is Localize Anything Model for 3D Medical images (MedLAM) \cite{lei2025medlsam}, which introduces unified anatomical mapping and multi-scale similarity learning. To estimate landmark localization, it applies Relative Position Regression (RPR) \cite{lei2021contrastive} followed by multi-scale feature similarity matching. Despite these advances, these methods are trained on datasets within the same modality and contrast voxels across augmented views of the same image, inherently limiting their generalization across heterogeneous imaging domains.

To enable landmark matching in cross-modality scenarios, UAE-M \cite{bai2025uae} extends the embedding-learning-based paradigm to CT-MRI settings via a two-stage learning process. At the first stage, a modality-agnostic network is trained on single-modality images with strong contrast augmentation. Subsequently, at the second stage, this modality-agnostic network initializes an iterative training process, which alternates between learning cross-modality embeddings from aligned multi-modality images and performing cross-modality registration to align those images. Despite its success, UAE-M relies on computationally expensive multi-stage iterative training to establish robust voxel correspondences between CT and MR images. By contrast, our method offers a simple, end-to-end trainable framework that learns domain-invariant voxel-wise embeddings from a distinct perspective by training on synthetic data generated via domain randomization.

\subsection{Domain randomization for medical image analysis}
Domain-randomization-based generative models aim to bridge all domain gaps via extreme augmentation beyond realism, eliminating the need for real training images \cite{tobin2017domain,tremblay2018training}. These methods fully randomize generation parameters to produce highly diverse training data, forcing networks to learn domain-independent features. For brain segmentation, methods like SynthSeg \cite{billot2023synthseg} use visually diverse pseudo-images synthesized from brain label maps to train networks, which can be applied to real brain MR scans across multiple domains without retraining. This paradigm has subsequently been extended to other segmentation tasks, including vascular structures \cite{chollet2024label}, cell nuclei \cite{dey2024anystar}, and abdominal organs \cite{krishnaswamy2025benchmarking}. For image registration, SynthMorph \cite{hoffmann2021synthmorph} proposed a generative strategy to produce diverse synthetic label maps and images, enabling networks to learn contrast-invariant registration. In the realm of representation learning, Anatomix \cite{dey2025learning} proposed a data engine that randomizes the spatial configuration of organ shape templates to create paired image samples with distinct appearances. These samples then train a network robust to nuisance imaging variations that do not alter image semantics. However, none of these works were designed for generalizable one-shot landmark localization. This task imposes a unique requirement: each voxel must be encoded with a discriminative embedding that sets it apart from other anatomical regions. Learning such universal discriminative voxel-wise embeddings is the core focus of our work, which is not directly addressed by existing domain-randomization-based methods.

\section{Method}

\subsection{Learning universal voxel-wise embedding}

\begin{figure*}[!t]
\centerline{\includegraphics[width=\textwidth]{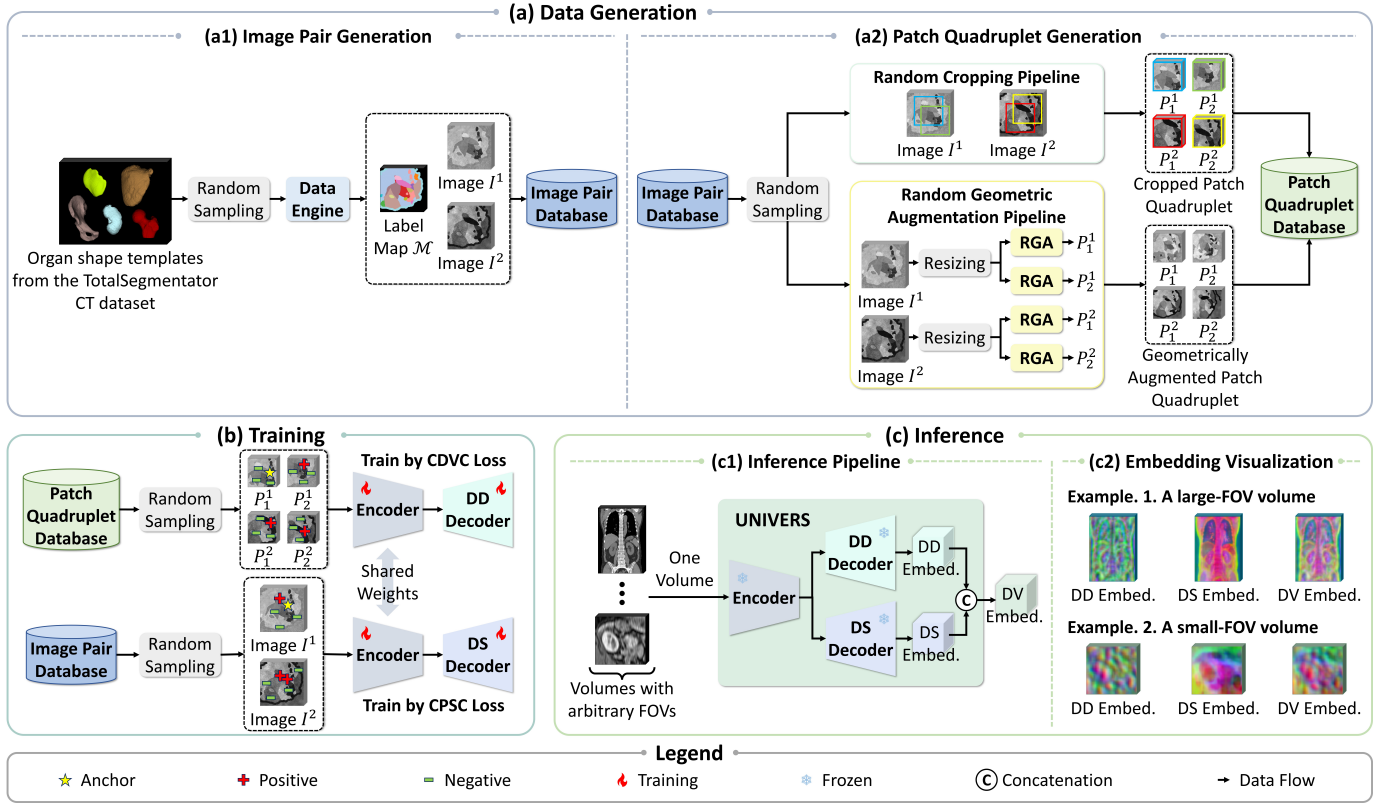}}
\caption{A schematic overview of UNIVERS. (a) We construct an image pair database and a patch quadruplet database from shape templates. (b) UNIVERS consists of a shared CNN encoder with a Domain-invariant Discriminative (DD) decoder and a Domain-invariant Semantic (DS) decoder. During training, patch quadruplets are randomly sampled to train the DD decoder via a Cross-domain Discriminative Voxel-wise Contrastive (CDVC) loss, while image pairs are randomly sampled to optimize the DS decoder via a Cross-domain Prototypical Semantic Contrastive (CPSC) loss. (c) During inference, UNIVERS can generate Domain-invariant Voxel-wise (DV) embeddings for multi-modality volumes with arbitrary FOVs. Embeddings are reduced to 3 dimensions via PCA and visualized as RGB images. \textbf{RGA}: Random Geometric Augmentation. \textbf{Embed.}: Embedding.}
\label{fig_overview}
\end{figure*}

UNIVERS learns domain-invariant voxel-wise embeddings without requiring any real training images. For training, we rely solely on freely available organ shape templates from the TotalSegmentator CT dataset \cite{wasserthal2023totalsegmentator}, which comprises 1,204 volumes with 104 annotated anatomical structures, to synthesize pseudo-training data. To encode precise location information for each voxel, UNIVERS is designed to acquire both high-level semantic understanding to distinguish different organs and low-level discriminative capacity to differentiate adjacent structures. To this end, we generate two specific types of pseudo-data, forming an image pair database and a patch quadruplet database (see Fig. \ref{fig_overview}-(a)). Each image pair consists of two pseudo-images that share a randomized semantic layout but differ in appearance, thereby facilitating domain-invariant semantic feature learning. In contrast, each patch quadruplet is derived from an image pair and comprises four overlapping pseudo-patches with distinct FOVs and spatial poses, encouraging local discriminability amidst varying contexts. UNIVERS then employs a dual-decoder network design: it comprises a shared Convolutional Neural Network (CNN) encoder with a Domain-invariant Semantic (DS) decoder and a Domain-invariant Discriminative (DD) decoder (see Fig. \ref{fig_overview}-(b)). The DS decoder is optimized on the image pairs by leveraging a Cross-domain Prototypical Semantic Contrastive (CPSC) loss, while the DD decoder is trained with the patch quadruplets and supervised via a Cross-domain Discriminative Voxel-wise Contrastive (CDVC) loss.

\subsubsection{Data generation}
\label{subsubsec_data_gen}
For training data, we first build the image pair database, based on which we further construct the patch quadruplet database.

\textbf{Image pair generation.} Randomized image pairs with rich semantic layouts and appearances are synthesized from shape templates via a data engine (see Fig. \ref{fig_overview}-(a1)). Following the pipeline introduced in \cite{dey2025learning}, the data engine first creates diverse randomized label maps with a fixed size of $S_I \times S_I \times S_I$ voxels using randomly sampled templates. For each randomized label map $\mathcal{M}$, it then synthesizes two pseudo-images, $I^1$ and $I^2$, which share an identical semantic layout but exhibit distinct appearances, effectively simulating aligned scans from two different domains. Such image synthesis involves sampling intensities from Gaussian Mixture Models (GMMs) with randomized parameters, followed by intensity and geometric augmentations. To preserve semantic alignment while maximizing appearance variance within an image pair, intensity augmentations are applied independently to $I^1$ and $I^2$, whereas geometric augmentations are shared. These pseudo-data constitutes the image pair database $\{(I^1,I^2|\mathcal{M})\}$.

\textbf{Patch quadruplet generation.} Randomized patch quadruplets with diverse FOVs and spatial poses are generated from the randomized image pairs. Specifically, to construct a patch quadruplet, we first randomly sample an image pair $(I^1,I^2)$, where $I^1$ and $I^2$ are spatially aligned. We then extract two patches from each image---$P_1^1$ and $P_2^1$ from $I^1$, and $P_1^2$ and $P_2^2$ from $I^2$. As shown in Fig. \ref{fig_overview}-(a2), we propose two distinct patch quadruplet generation pipelines that separately simulate different sources of spatial variation, and randomly select one per quadruplet:
\begin{itemize}
\item \textbf{Random cropping pipeline:} This pipeline predominantly simulates patches with large FOV variations. Each patch is randomly cropped from an image at a random location with a random size of $S_R\times S_R\times S_R$ voxels, where $S_R$ is sampled from a uniform distribution $U(S_L, S_U)$. Here $S_L$ and $S_U$ respectively denote the lower bound and the upper bound of $S_R$, which are constrained to ensure that the four cropped patches spatially overlap (see Section \ref{subsec_implement_detail} for detailed parameter values). All patches are subsequently resized to $S_P\times S_P\times S_P$ voxels. 
\item \textbf{Random geometric augmentation pipeline:} This pipeline primarily simulates patches with large spatial pose variations. Each image is first resized to $S_P\times S_P\times S_P$ voxels and then duplicated. The duplicated patches then independently undergo random geometric augmentations, including rotation, translation, and scaling.
\end{itemize}
This random selection strategy decouples the sources of spatial variation, enabling more stable and effective optimization of the network. In contrast, applying large FOV shifts and severe geometric deformations simultaneously may introduce overly destructive variations, making it difficult for the network to establish reliable voxel correspondences during contrastive learning. By separating individual variation sources within each quadruplet, the network can more effectively learn discriminative embeddings. Finally, these pseudo-data form the patch quadruplet database $\{(P_1^1, P_2^1, P_1^2, P_2^2)\}$.

\subsubsection{Training}

Built on the shared CNN encoder, the DS decoder and the DD decoder are optimized with distinct training objectives detailed below.

\textbf{Domain-invariant Semantic (DS) decoder.} This decoder is designed to learn domain-invariant semantic voxel-wise embeddings from the image pair database. Guided by semantic label maps, we supervise the learning process with a Cross-domain Prototypical Semantic Contrastive (CPSC) loss. Unlike previous works that rely on single-modality inputs for supervised contrastive learning \cite{khosla2020supervised}, our method leverages cross-domain pseudo-image pairs, thereby enforcing the alignment of semantic embeddings across diverse domains. As illustrated in Fig. \ref{fig_overview}-(b), given a randomly sampled image pair $(I^1,I^2|\mathcal{M})$, we pass them through the shared encoder and the DS decoder to extract the DS embeddings. The CPSC loss $\mathcal{L}_{CPSC}$ is then computed as:
\begin{equation}
\begin{aligned}
\mathcal{L}_{CPSC} &= - \sum_{c=1}^{N^C} \frac{1}{N_c^V} \sum_{v=1}^{N_c^V} \\
&\mathrm{log} \bigg ( \frac{\mathrm{exp}(f_{c}^S\cdot f_{c,v}^{S} / \tau_S)}{\sum_{c'=1}^{N^C} \sum_{v'=1}^{N_{c'}^V} \mathrm{exp}(f_{c}^S\cdot f_{c',v'}^{S} / \tau_S)} \bigg )
\label{eq_CPSC}
\end{aligned}
\end{equation}
where $N^C$ is the number of classes in the label map $\mathcal{M}$; $N_c^V$ and $N_{c'}^V$ denote the number of voxels belonging to classes $c$ and $c'$ across the image pairs, respectively. $f_{c,v}^{S}$ and $f_{c',v'}^{S}$ represent the $\ell_2$-normalized DS embeddings of voxels $v$ and $v'$ for classes $c$ and $c'$, respectively. $f_{c}^S=\frac{1}{N_c^V}\sum_{v=1}^{N_c^V} f_{c,v}^{S}$ acts as the cross-domain prototype for class $c$. $\tau_S$ is the temperature parameter for the CPSC loss, and ``$\cdot$'' represents the inner product. By pulling voxel-wise embeddings toward their corresponding cross-domain class prototypes while pushing them away from other class prototypes, this loss effectively constructs a domain-invariant semantic space that facilitates organ differentiation in multi-domain data.

\textbf{Domain-invariant Discriminative (DD) decoder.} This decoder learns domain-invariant discriminative voxel-wise embeddings using the patch quadruplet database. To this end, we propose a Cross-domain Discriminative Voxel-wise Contrastive (CDVC) loss. Different from the single-positive voxel-wise contrastive loss introduced in \cite{yan2022sam} that focuses on single-modality patch pairs, our CDVC loss operates on multi-domain pseudo-patch quadruplets and employs a multi-positive contrastive strategy, thereby effectively establishing cross-domain voxel-wise correspondences. In particular, a patch quadruplet $(P_1^1, P_2^1, P_1^2, P_2^2)$ is first randomly sampled and then fed into the shared encoder and the DD decoder to extract the DD embeddings. For each patch $P_j^i$ ($i,j\in\{1,2\}$), we compute its overlapping region with the other three patches, i.e., the set of voxels that have corresponding voxels in each of the other three patches. Within this region, we randomly sample $N^{A}$ voxels as anchors from each patch, yielding a total of $4 N^{A}$ anchor voxels across the four patches. For each anchor voxel in patch $P_j^i$, we identify $N^{P}=3$ positive voxels: one from each of the other three patches that maps to the same spatial location in the original image volume (see Fig. \ref{fig_overview}-(b) for an illustration). Additionally, $N^{N}$ negative voxels per anchor are selected from all four patches following the hard and diverse negative sampling strategy introduced in \cite{yan2022sam}. The CDVC loss $\mathcal{L}_{CDVC}$ is formulated as:
\begin{equation}
\begin{aligned}
&\mathcal{L}_{CDVC} = -\frac{1}{4 N^A N^P} \sum_{a=1}^{4N^A} \bigg ( \sum_{p=1}^{N^P}\\
&  \mathrm{log} \bigg ( \frac{\mathrm{exp}(f_{a}^D\cdot f_{a,p}^{D+} / \tau_{D})}{\mathrm{exp}(f_{a}^D\cdot f_{a,p}^{D+} / \tau_{D}) + \sum_{n=1}^{N^N} \mathrm{exp}(f_{a}^D\cdot f_{a,n}^{D-} / \tau_{D})} \bigg ) \bigg )
\end{aligned}
\label{eq_voxel_contrast}
\end{equation}
where $f_{a}^D$, $f_{a,p}^{D+}$, and $f_{a,n}^{D-}$ denote the $\ell_2$-normalized DD embeddings of anchor voxel $a$, its positive voxel $p$, and its negative voxel $n$, respectively. $\tau_{D}$ is the temperature parameter for the CDVC loss. By employing this multi-positive contrastive mechanism, the CDVC loss simultaneously enforces feature affinity across multi-domain positives while repelling negative samples, thereby promoting robust domain-invariant and discriminative representations of anatomical structures.

The DS decoder and the DD decoder are trained by joint optimization. The overall training loss function $\mathcal{L}$ is a weighted sum of the CPSC loss and the CDVC loss:
\begin{equation}
\mathcal{L}=\mathcal{L}_{CPSC}+\lambda\mathcal{L}_{CDVC}
\label{eq_loss}
\end{equation}
where $\lambda$ is a weighting factor that balances the two loss terms.

\subsubsection{Inference}
Without requiring any retraining or finetuning, UNIVERS generalizes to diverse unseen real clinical domains during inference. As depicted in Fig. \ref{fig_overview}-(c), given an arbitrary FOV volume, the final Domain-invariant Voxel-wise (DV) embedding is produced by concatenating the outputs of the DS decoder and the DD decoder, followed by $\ell_2$-normalization.

\begin{figure*}[!t]
\centerline{\includegraphics[width=\textwidth]{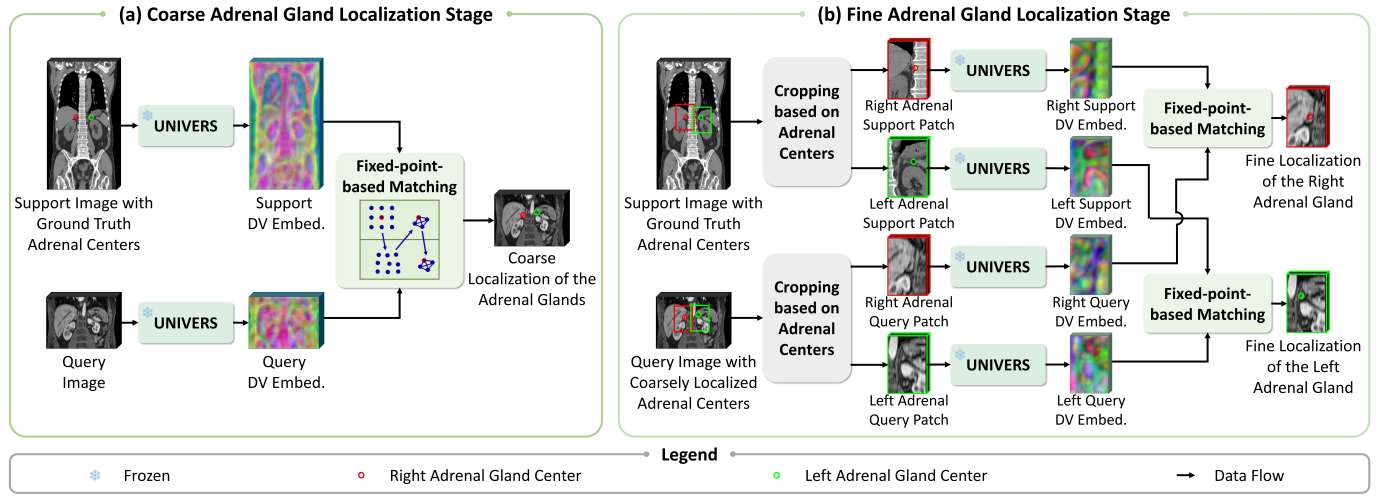}}
\caption{Illustration of the coarse-to-fine adrenal gland localization framework built upon a single frozen UNIVERS network. Adrenal gland centers are shown as projections of 3D points along the frontal plane for visualization purposes only. Embeddings are reduced to 3 dimensions via PCA and visualized as RGB images. Embed.: Embedding.}
\label{fig_localization_framework}
\end{figure*}

\subsection{Application to landmark localization}

\subsubsection{One-shot landmark localization}
\label{subsubsec_oneshot_loc}
UNIVERS naturally extends to one-shot landmark localization. Given a single support image with a labeled landmark and an unlabeled query image, which may be acquired from a different modality and/or with a different FOV, we first extract the DV embeddings for both images. Trained with our proposed learning strategy, the DV embedding space ensures that corresponding anatomical landmarks across varying domains are mapped to similar representations, while non-corresponding landmarks are pushed apart. Exploiting this domain-invariant voxel-wise discriminability, the corresponding landmark in the query image can be localized using a fixed-point-based matching method as introduced in \cite{bai2025uae}.

\subsubsection{Coarse-to-fine one-shot adrenal gland localization}
To validate the effectiveness of UNIVERS, we apply it to adrenal gland localization. Due to the small size of the adrenal glands within arbitrary FOV scans, we propose a coarse-to-fine one-shot framework for precise localization, as illustrated in Fig. \ref{fig_localization_framework}. A single frozen UNIVERS network is used for both the coarse and fine localization stages, benefiting from its capability to handle volumes with arbitrary FOVs. Given a support image with labeled left and right adrenal centers and an unlabeled query image, the localization proceeds in two steps:

\textbf{Coarse localization.} The support and query images are first resampled to an isotropic spacing of $R_C\times R_C\times R_C~{\rm mm}^3$. One-shot landmark localization is then performed to propagate the labeled support centers to the query image.
 
\textbf{Fine localization.} To achieve higher precision, the images are resampled to a finer isotropic spacing of $R_F\times R_F\times R_F~{\rm mm}^3$ ($R_F<R_C$). We then crop bilateral patches (size: $X\times Y \times Z$ voxels) centered at the labeled support centers and the coarsely estimated query centers, respectively. The cropping boundaries are constrained to prevent overlap between the left and right patches and to avoid exceeding the image boundaries. After mapping the support center coordinates into the local patch coordinate systems, we independently apply one-shot landmark localization to each left and right patch pair. Finally, the refined center coordinates in the query patches are mapped back into the original query image coordinate system.

\subsection{Implementation details}
\label{subsec_implement_detail}
We implemented the proposed UNIVERS framework using PyTorch v2.1. For data generation, we synthesized 120,000 randomized image pairs during offline construction. To further enhance data diversity, online augmentations were applied, including Gaussian noise, Gamma shifts, Gaussian blurring, elastic deformations, resolution degradations, affine transformations, and axis flipping. Patch quadruplets were generated online from randomly sampled image pairs. The size parameters used for image and patch generation were empirically set to $S_I=128$, $S_L=77$, $S_U=115$, and $S_P=96$ voxels.

For training, we empirically set $N_A=100$, $N_N=500$, $\tau_S=0.1$, $\tau_{D}=0.5$, and $\lambda=1.0$. The DS and DD decoders processed image pairs and patch quadruplets with batch sizes of 1 and 2, respectively. The network was optimized using Stochastic Gradient Descent (SGD) with a momentum of 0.9 and a weight decay of $1.0 \times 10^{-4}$. An initial learning rate of 0.02 was adopted, scheduled with warmup and step decay. Training lasted for 50,000 iterations.

For the coarse-to-fine adrenal gland localization framework, the resampled isotropic spacings for the coarse and fine stages were $R_C=2.0$ mm and $R_F=1.0$ mm, respectively. During fine localization, the images were cropped to a size of $X=94$, $Y=125$ and $Z=125$ voxels.

\section{Experiments and results}

\subsection{Datasets and evaluation settings}
\label{subsec_dataset_setting}
To comprehensively evaluate the capability of UNIVERS on generalizable adrenal gland localization, we conducted experiments under three one-shot settings with increasing difficulty: intra-dataset, cross-dataset intra-modality, and cross-dataset cross-modality. Three datasets were employed:
\subsubsection{MAGI} The Multi-field-of-view Adrenal Gland Image (MAGI) dataset consists of 644 CT images retrospectively collected from Shanghai General Hospital after ethical approval (approved by Shanghai General Hospital Institutional Review Board with approval no. 2026KS557. Informed consent was waived.). Image sizes range from $512\times512\times51$ to $512\times512\times1168$ voxels, with spacing ranging from $0.55\times 0.55\times 0.63$ to $0.98\times 0.98\times 1.50~{\rm mm}^3$. For all CT images, adrenal gland regions were annotated by a committee of clinical experts, comprising two senior surgeons and four junior residents. The centroids of the annotated masks were then computed and adopted as the ground truth (GT) adrenal gland center points. We randomly split the data into 480 training, 52 validation, and 112 test cases.
\subsubsection{DAP Atlas} The Dense Anatomical Prediction (DAP) Atlas dataset \cite{jaus2023towards} provides anatomical masks for 533 whole-body CT scans from AutoPET \cite{gatidis2022whole}. We extracted the adrenal gland masks from the annotations and calculated their centroids as GT adrenal center points. The data were randomly partitioned into 360 training, 41 validation, and 132 test cases.
\subsubsection{MRISegmentator} This abdominal MRI segmentation dataset \cite{zhuang2024mrisegmentator} comprises 780 multi-sequence scans. Among these, 729 scans containing both the left and right adrenal glands were used in our experiment. We obtained the GT adrenal gland center points from the adrenal gland masks. The data were partitioned according to the official split, yielding 454 scans for training, 44 for validation, and 231 for testing.

\begin{figure}[!t]
\centerline{\includegraphics[width=\columnwidth]{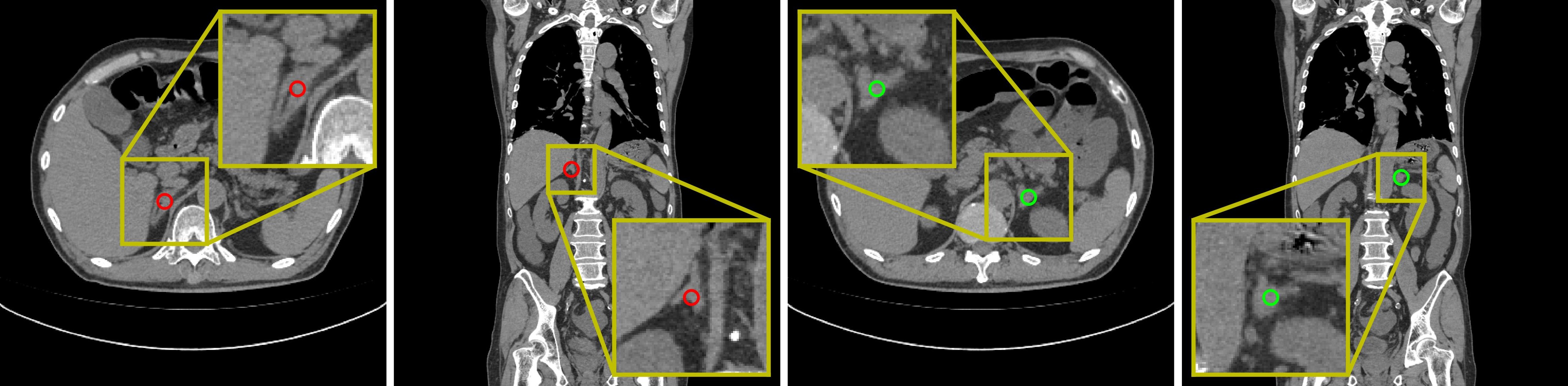}}
\caption{Visualization of the single support sample for all three one-shot settings. The left two columns respectively display transverse and frontal slices passing through the GT right adrenal gland center points (red circles), while the right two columns show corresponding slices for the left adrenal gland center points (green circles).}
\label{fig_support}
\end{figure}

\textbf{Evaluation settings.} For all three one-shot settings, a single, identical support sample from the training set of the MAGI dataset was adopted (see Fig. \ref{fig_support}). Query test samples for the three settings were from three different sources: the test set of the MAGI dataset (intra-dataset), the test set of the DAP Atlas dataset (cross-dataset intra-modality), and the test set of the MRISegmentator dataset (cross-dataset cross-modality), respectively. Except for this single support sample, the training and validation sets of all datasets were strictly excluded from use by any one-shot methods. They were only utilized during the training of the upper bound models (see Section~\ref{subsec_comparison_method}).

\textbf{Evaluation metrics.} We adopted the Mean Euclidean Distance (MED) and the Success Detection Rate (SDR) \cite{huang2024landmark} as the evaluation metrics. MED is the average Euclidean distance between the predicted and GT adrenal gland center points. SDR measures the percentage of predictions within a predefined distance threshold. We reported SDR at thresholds of 10 mm and 20 mm (referred to as ${\rm SDR}_{\rm 10}$ and ${\rm SDR}_{\rm 20}$, respectively). To compare different competing methods, we conducted Wilcoxon signed-rank test \cite{woolson2007wilcoxon} and chose a significant level of 0.05. 

\subsection{Comparison methods}
\label{subsec_comparison_method}
We compared UNIVERS against SOTA one-shot landmark localization methods, including both registration-based and embedding-learning-based approaches. Registration-based methods included Affine \cite{marstal2016simpleelastix} and ConvexAdam \cite{siebert2024convexadam}. Embedding-learning-based methods covered MedLAM \cite{lei2025medlsam}, SAM \cite{yan2022sam}, UAE-S \cite{bai2025uae} and UAE-M \cite{bai2025uae}. We additionally included Anatomix \cite{dey2025learning}, which generates pseudo-training data using the same shape templates as our work and learns general-purpose semantic representations. Although its training objective does not optimize for discriminative embeddings, we used it as a synthetic-data-based representation learning baseline. 

In addition to the above competing methods, we established two reference baselines for each evaluation setting. These models were developed using nnLandmark \cite{ertl2026nnlandmark}, a SOTA supervised medical landmark localization framework. The first model, referred to as the upper bound model, was trained on the full training set of the same dataset from which the test samples were drawn. The second model, denoted as the lower bound model, was trained solely on the single support sample specified in Section \ref{subsec_dataset_setting}, mimicking the data availability constraint of the one-shot methods.

\subsection{Comparison results}
\label{subsec_comparison_results}

\begin{figure}[!t]
\centerline{\includegraphics[width=\columnwidth]{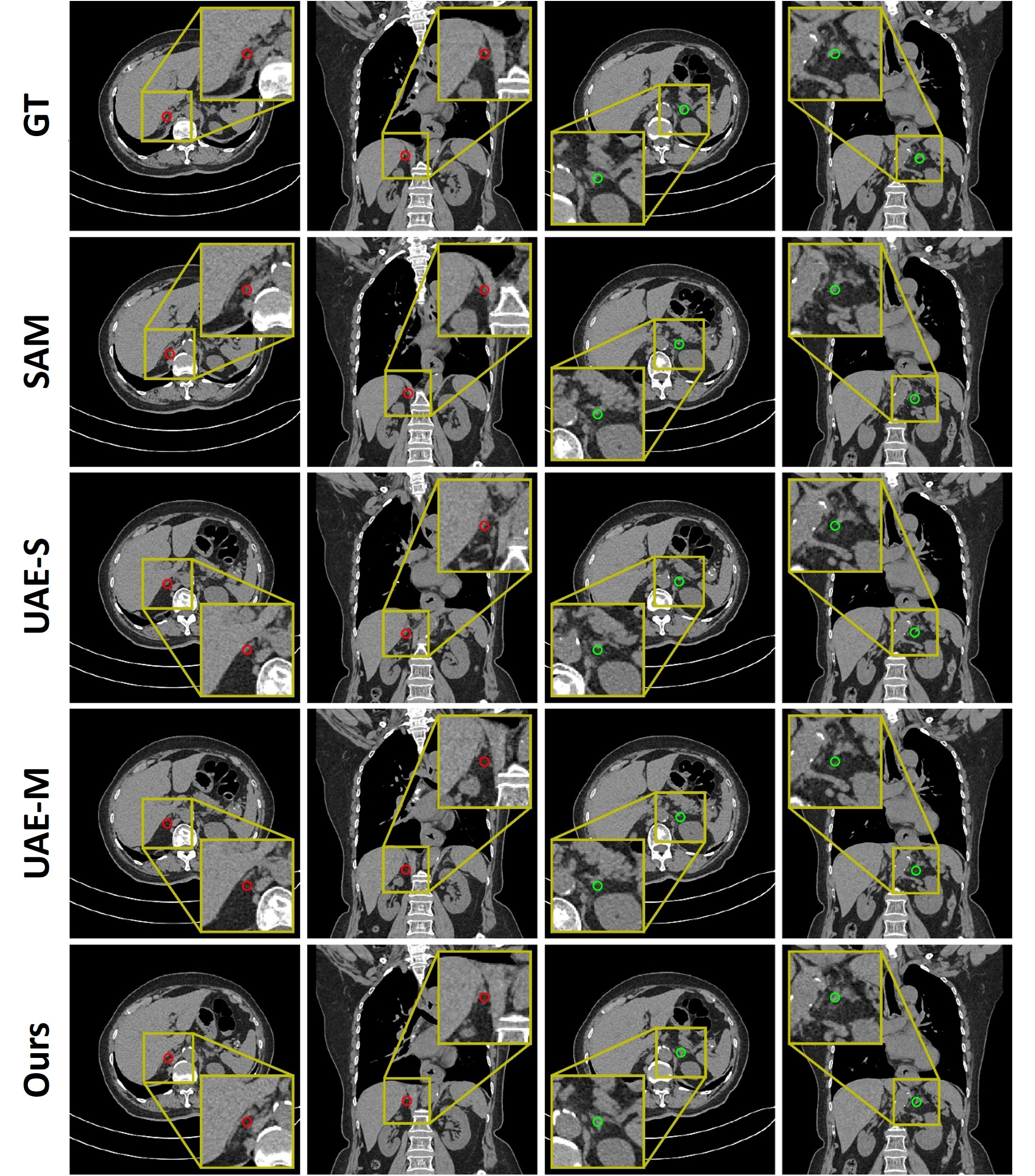}}
\caption{Visual comparison of intra-dataset one-shot adrenal gland localization. The left two columns respectively display transverse and frontal slices passing through the GT or predicted right adrenal gland center points (red circles), while the right two columns show corresponding slices for the left adrenal gland center points (green circles).}
\label{fig_compare_CT}
\end{figure}

\begin{table}[!tbp]
\centering
\caption{Quantitative comparison of intra-dataset one-shot adrenal gland localization. Bold indicates the best results, and underline indicates the second-best results. $\uparrow$ indicates higher is better, $\downarrow$ indicates lower is better.}
\begin{tabular}{lcccc}
\toprule
\multirow{2}{*}{Method} & ${\rm SDR}_{\rm 10}$  & ${\rm SDR}_{\rm 20}$ & ${\rm MED}$ & \multirow{2}{*}{$p$-value} \\ 
& (\%) $\uparrow$ & (\%) $\uparrow$ &  (mm) $\downarrow$ & \\
\midrule
\multicolumn{5}{l}{\textbf{\textit{Reference Baselines}}} \\
Lower Bound & 55.80 & 75.45  & 31.63$\pm$65.29 & -\\
Upper Bound & 98.66 & 99.55  & 3.04$\pm$13.26 & -\\
\midrule
\multicolumn{5}{l}{\textbf{\textit{One-shot Methods}}} \\
Affine \cite{marstal2016simpleelastix} & 4.91 & 36.16  & 35.37$\pm$31.19 & $<$0.001\\
ConvexAdam \cite{siebert2024convexadam} & 46.43 & 82.14 & 18.82$\pm$28.12 & $<$0.001\\
Anatomix \cite{dey2025learning} & 0.45 & 3.57 & 115.73$\pm$60.96 & $<$0.001\\
MedLAM \cite{lei2025medlsam} & 31.25 & 72.32  & 17.30$\pm$13.38 & $<$0.001\\
SAM \cite{yan2022sam} & 64.29 & 90.63 &  9.91$\pm$7.46  & $<$0.001 \\
UAE-S \cite{bai2025uae} & \underline{73.66} & \textbf{99.55}  & \underline{7.65$\pm$3.80} & $<$0.001\\
UAE-M \cite{bai2025uae} & 65.18 & 94.20 & 9.14$\pm$5.39 & $<$0.001\\
UNIVERS (Ours) &  \textbf{87.50} & \textbf{99.55} & \textbf{6.16$\pm$3.73} & -\\
\bottomrule
\end{tabular}
\label{tab_compare_adrenal_detect_CT}
\end{table}

\subsubsection{Results under the intra-dataset one-shot setting}
Table \ref{tab_compare_adrenal_detect_CT} reports the quantitative results under the intra-dataset one-shot evaluation setting. We first examined the reference baselines. Specifically, the upper bound model achieved an MED of 3.04 mm and an ${\rm SDR}_{\rm 10}$ of 98.66\%, while the lower bound model attained an MED of 31.63 mm and an ${\rm SDR}_{\rm 10}$ of 55.80\%. These results reveal substantial performance degradation when models are trained by extremely scarce annotated data. We then compared our proposed UNIVERS with the one-shot competing methods. As shown in the table, one can see that the proposed approach outperformed all competing methods, achieving the lowest MED (6.16 mm) and the highest ${\rm SDR}_{\rm 10}$ (87.50\%). In contrast, the second-best performing method (UAE-S \cite{bai2025uae}) yielded an MED of 7.65 mm and an ${\rm SDR}_{\rm 10}$ of 73.66\%. Wilcoxon signed-rank test \cite{woolson2007wilcoxon} on the MED results of UNIVERS and all one-shot comparison methods yielded $p$-values smaller than 0.001, indicating statistically significant improvements. Fig. \ref{fig_compare_CT} provides a visual comparison between UNIVERS and the top-three one-shot competing methods. It can be observed that our method delivers the most precise adrenal localization among all one-shot competing approaches.

\begin{figure}[!t]
\centerline{\includegraphics[width=\columnwidth]{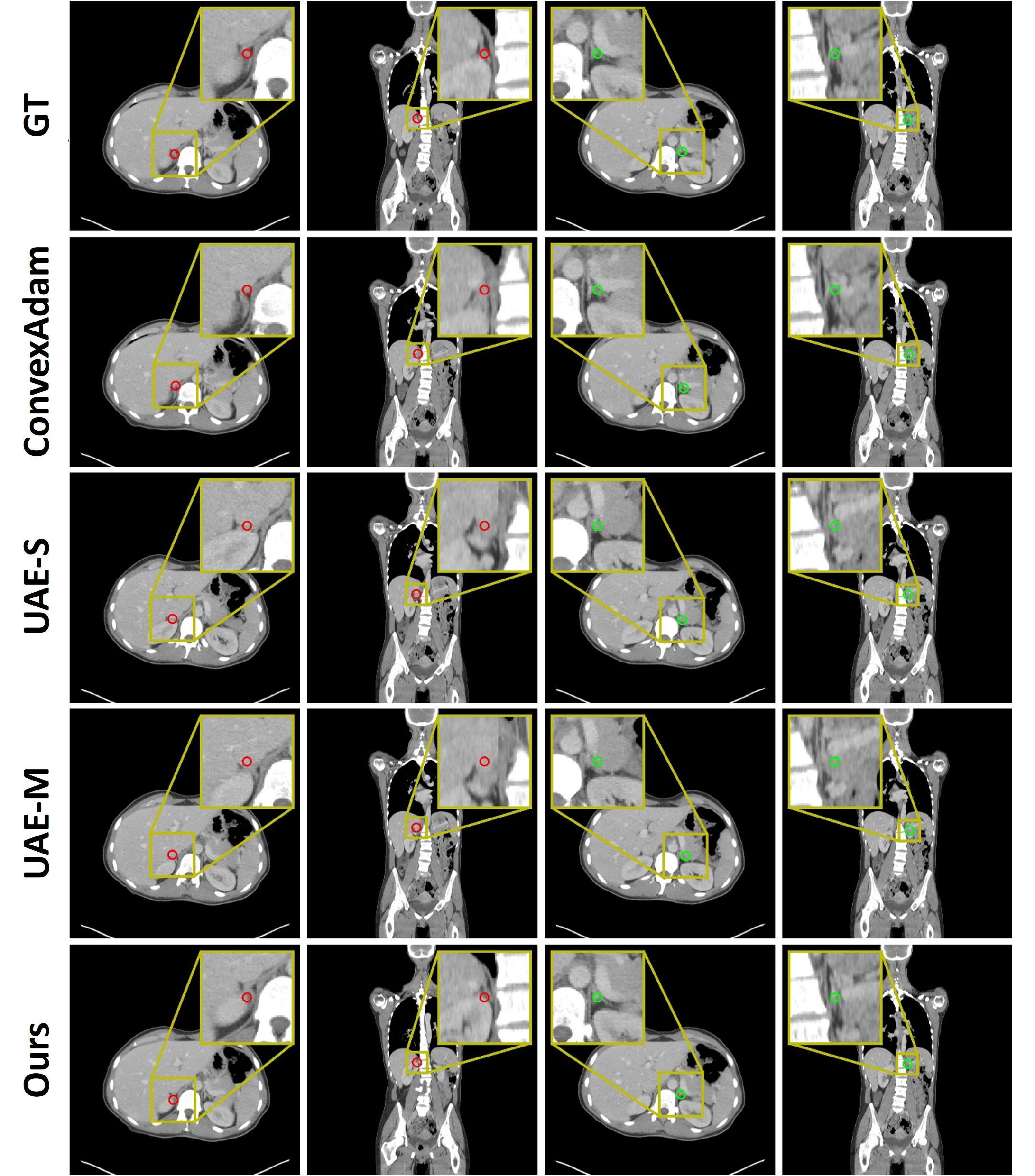}}
\caption{Visual comparison of cross-dataset intra-modality one-shot adrenal gland localization. The left two columns respectively display transverse and frontal slices passing through the GT or predicted right adrenal gland center points (red circles), while the right two columns show those slices for the left adrenal gland center points (green circles).}
\label{fig_compare_CT_DAP}
\end{figure}

\begin{table}[!tbp]
\centering
\caption{Quantitative comparison of cross-dataset intra-modality one-shot adrenal gland localization. Bold indicates the best results, and underline indicates the second-best results. $\uparrow$ indicates higher is better, $\downarrow$ indicates lower is better.}
\begin{tabular}{lcccc}
\toprule
\multirow{2}{*}{Method} & ${\rm SDR}_{\rm 10}$  & ${\rm SDR}_{\rm 20}$ & ${\rm MED}$ & \multirow{2}{*}{$p$-value} \\ 
& (\%) $\uparrow$ & (\%) $\uparrow$ &  (mm) $\downarrow$ & \\
\midrule
\multicolumn{5}{l}{\textbf{\textit{Reference Baselines}}} \\
Lower Bound & 0.00 & 0.00 & 351.56$\pm$106.53 & -\\
Upper Bound &  96.59 & 98.86 & 3.70$\pm$3.36 & -\\
\midrule
\multicolumn{5}{l}{\textbf{\textit{One-shot Methods}}} \\
Affine \cite{marstal2016simpleelastix} & 9.85 & 43.18 & 26.46$\pm$20.89 & $<$0.001 \\
ConvexAdam \cite{siebert2024convexadam} & 42.42 & 85.61 & 15.04$\pm$16.10 & $<$0.001 \\
Anatomix \cite{dey2025learning} & 0.00 & 0.76 & 100.59$\pm$48.16 & $<$0.001 \\
MedLAM \cite{lei2025medlsam} & 10.98 & 45.45 & 28.35$\pm$20.10  & $<$0.001 \\
SAM \cite{yan2022sam} &  33.71 & 68.56 & 17.09$\pm$26.22 & $<$0.001  \\
UAE-S \cite{bai2025uae} &  \underline{70.45} & 94.70 & \underline{8.34$\pm$5.73} & 0.047 \\
UAE-M \cite{bai2025uae} &  57.58 & \underline{95.45} & 10.03$\pm$5.33 & $<$0.001\\
UNIVERS (Ours) &  \textbf{78.03} & \textbf{98.48} & \textbf{7.46$\pm$4.50} & -\\
\bottomrule
\end{tabular}
\label{tab_compare_adrenal_detect_CT_DAP}
\end{table}

\subsubsection{Results under the cross-dataset intra-modality one-shot setting}
Table \ref{tab_compare_adrenal_detect_CT_DAP} summarizes the quantitative results under the cross-dataset intra-modality one-shot setting. We began by investigating the performance of the reference baselines. It showed a sharp performance drop: the upper bound model achieved an MED of 3.70 mm and an ${\rm SDR}_{\rm 10}$ of 96.59\%, while the lower bound model recorded an MED of 351.56 mm and an ${\rm SDR}_{\rm 10}$ of 0.00\%. This dramatic decline underscores the dual challenge of data scarcity and domain shift across datasets. We further evaluated the performance of the proposed UNIVERS and the one-shot competing methods. From Table \ref{tab_compare_adrenal_detect_CT_DAP}, one can see that UNIVERS achieved superior generalizable one-shot localization performance, with an MED of 7.46 mm and an ${\rm SDR}_{\rm 10}$ of 78.03\%. In contrast, the second-best performing method (UAE-S \cite{bai2025uae}) obtained an MED of 8.34 mm and an ${\rm SDR}_{\rm 10}$ of 70.45\%. To compare UNIVERS with the one-shot competing methods, we further performed Wilcoxon signed-rank test \cite{woolson2007wilcoxon} on the MED results. All obtained $p$-values were less than 0.05, indicating that the improvements achieved by our method were statistically significant. In Fig. \ref{fig_compare_CT_DAP}, we visually compare the localization performance of UNIVERS with the top-three one-shot competing methods. It can be observed that UNIVERS produces the most accurate landmark localization.

\begin{figure}[!t]
\centerline{\includegraphics[width=\columnwidth]{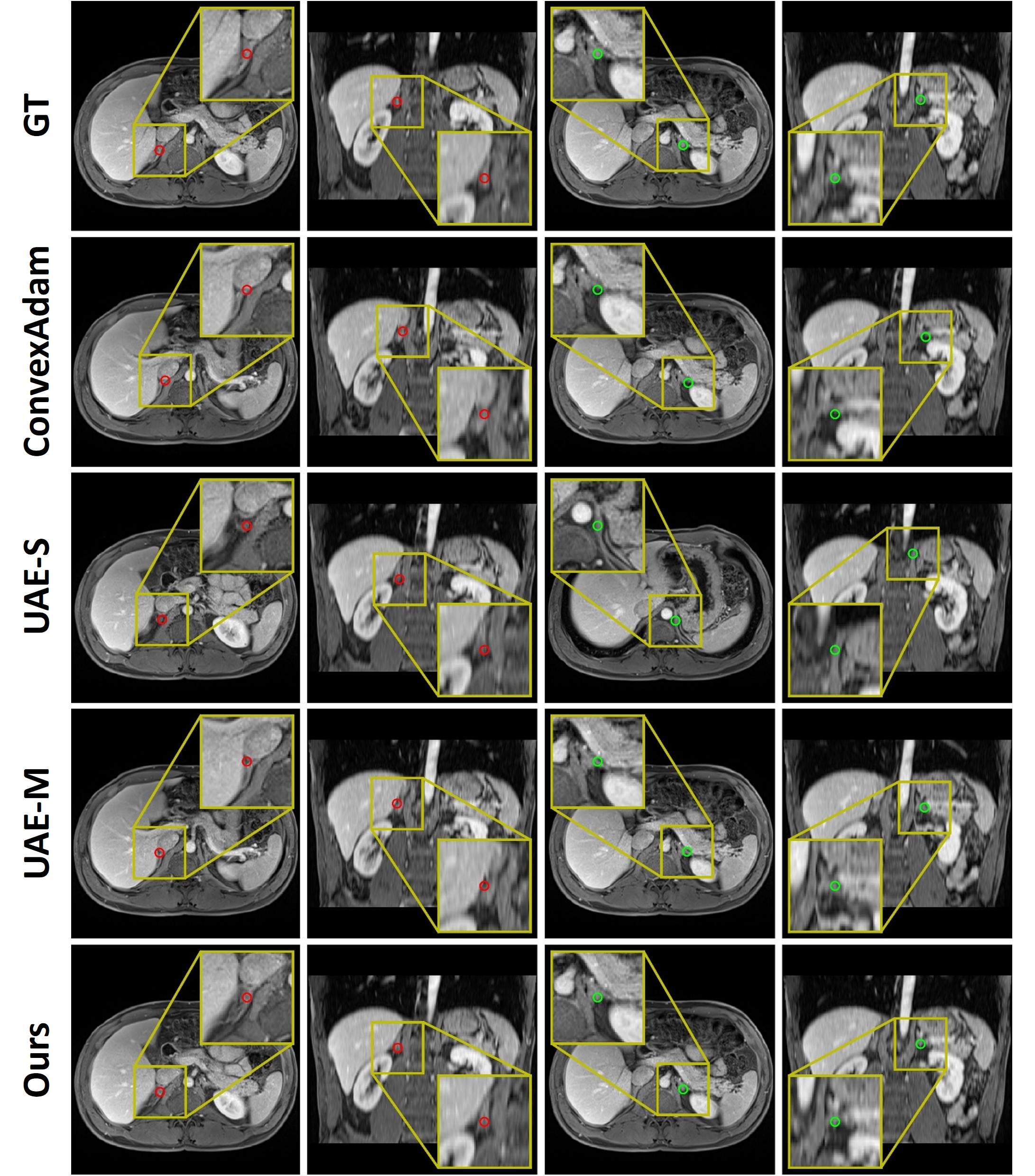}}
\caption{Visual comparison of cross-dataset cross-modality one-shot adrenal gland localization. The left two columns respectively display transverse and frontal slices passing through the GT or predicted right adrenal gland center points (red circles), while the right two columns show those slices for the left adrenal gland center points (green circles).}
\label{fig_compare_MR}
\end{figure}

\begin{table}[!tbp]
\centering
\caption{Quantitative comparison of cross-dataset cross-modality one-shot adrenal gland localization. Bold indicates the best results, and underline indicates the second-best results. $\uparrow$ indicates higher is better, $\downarrow$ indicates lower is better.}
\begin{tabular}{lcccc}
\toprule
\multirow{2}{*}{Method} & ${\rm SDR}_{\rm 10}$  & ${\rm SDR}_{\rm 20}$ & ${\rm MED}$ & \multirow{2}{*}{$p$-value} \\ 
& (\%) $\uparrow$ & (\%) $\uparrow$ &  (mm) $\downarrow$ & \\
\midrule
\multicolumn{5}{l}{\textbf{\textit{Reference Baselines}}} \\
Lower Bound & 0.00 & 0.22 & 120.70$\pm$46.76 & - \\
Upper Bound & 88.96 & 97.19 & 5.56$\pm$6.05 & -\\
\midrule
\multicolumn{5}{l}{\textbf{\textit{One-shot Methods}}} \\
Affine \cite{marstal2016simpleelastix} & 3.46 & 26.84 & 30.30$\pm$15.48  & $<$0.001 \\
ConvexAdam \cite{siebert2024convexadam} & 29.65 & 76.84 & 16.52$\pm$11.82 & $<$0.001  \\
Anatomix \cite{dey2025learning} & 0.00 & 2.16 & 56.68$\pm$18.62  & $<$0.001 \\
MedLAM \cite{lei2025medlsam} & 2.16 & 7.36 & 74.89$\pm$35.33  & $<$0.001 \\
SAM \cite{yan2022sam} &  8.23 & 24.03 & 82.37$\pm$78.63  & $<$0.001  \\
UAE-S \cite{bai2025uae} &  5.63 & 25.76 & 49.54$\pm$47.02  & $<$0.001  \\
UAE-M \cite{bai2025uae} & \underline{41.77} & \underline{90.69} & \underline{12.29$\pm$6.95}  & $<$0.001\\
UNIVERS (Ours) & \textbf{64.72} & \textbf{91.77} & \textbf{9.65$\pm$7.61} & - \\
\bottomrule
\end{tabular}
\label{tab_compare_adrenal_detect_MR}
\end{table}

\subsubsection{Results under the cross-dataset cross-modality one-shot setting}
Table \ref{tab_compare_adrenal_detect_MR} presents the experimental results under the cross-dataset cross-modality one-shot setting. We first investigated the performance of the reference baselines. As reported in this table, the lower bound model performed poorly, yielding an MED of 120.70 mm and an ${\rm SDR}_{\rm 10}$ of 0.00\%. In contrast, the upper bound model achieved an MED of 5.56 mm and an ${\rm SDR}_{\rm 10}$ of 88.96\%. Such a large gap highlights the difficulty of cross-modality one-shot localization, where both data scarcity and cross-modality domain shifts are present. We then compared UNIVERS with the one-shot competing methods. From Table \ref{tab_compare_adrenal_detect_MR}, one can see that our method demonstrated superior cross-modality generalization performance. In particular, our method achieved an MED of 9.65 mm, yielding a reduction of 2.64 mm compared with the second-best performing method (UAE-M \cite{bai2025uae}: 12.29 mm). Meanwhile, our approach attained an ${\rm SDR}_{\rm 10}$ of 64.72\%, exceeding the second-best performing method (UAE-M \cite{bai2025uae}: 41.77\%) by a large margin of 22.95\%. We further conducted Wilcoxon signed-rank test \cite{woolson2007wilcoxon} on the MED results to compare UNIVERS with the one-shot competing methods. The obtained $p$-values were all smaller than 0.001, indicating statistically significant differences. Fig. \ref{fig_compare_MR} visually compares the localization results achieved by the proposed method and the top-three one-shot competing methods. One can see from this figure that our method produces the most accurate adrenal centers.

\subsection{Ablation study}

\begin{table*}[!tbp]
\centering
\caption{Quantitative results of the ablation study on all three one-shot settings. Bold indicates the best results. $\uparrow$ indicates higher is better, $\downarrow$ indicates lower is better. "w/o": without.}
\begin{tabular}{ccccccc}
\toprule
\multirow{2}{*}{Model Configuration}  & \multicolumn{2}{c}{Intra-dataset} & \multicolumn{2}{c}{Cross-dataset Intra-modality} & \multicolumn{2}{c}{Cross-dataset Cross-modality}  \\ 
\cmidrule(lr){2-3} 
\cmidrule(lr){4-5} 
\cmidrule(lr){6-7} 
  & ${\rm SDR}_{\rm 10}$ (\%) $\uparrow$  & ${\rm MED}$ (mm) $\downarrow$ &  ${\rm SDR}_{\rm 10}$  (\%) $\uparrow$ & ${\rm MED}$ (mm) $\downarrow$  &  ${\rm SDR}_{\rm 10}$ (\%) $\uparrow$ & ${\rm MED}$ (mm) $\downarrow$   \\ 
\midrule
Full Model (Ours) &  \textbf{87.50} & \textbf{6.16$\pm$3.73} & \textbf{78.03} & \textbf{7.46$\pm$4.50} & \textbf{64.72} & \textbf{9.65$\pm$7.61} \\
\midrule
\multicolumn{7}{l}{\textbf{(a) Ablation on the Pseudo-patch Quadruplet Generation Strategy}} \\
\midrule
Cropping-only &  76.34 & 8.20$\pm$7.43 & 61.74 & 11.35$\pm$11.20 & 58.44 & 12.34$\pm$11.88 \\
Geometric-augmentation-only & 83.48 & 7.38$\pm$8.17 & 61.74 & 12.15$\pm$15.12 & 54.98 & 11.03$\pm$8.44 \\
Cropping-and-geometric-augmentation & 85.71 & 6.64$\pm$5.59 & 68.56 & 9.70$\pm$9.68 & 55.84 & 11.41$\pm$9.12 \\
\midrule
\multicolumn{7}{l}{\textbf{(b) Ablation on the contrastive learning strategy}} \\
\midrule
w/o CPSC &  65.63 & 10.00$\pm$10.40 & 54.17 & 13.71$\pm$21.27 & 50.00 & 13.22$\pm$10.80 \\
Intra-domain Single-positive CDVC &  83.93 & 7.05$\pm$7.67 & 59.85 & 20.21$\pm$32.34 & 27.06 & 24.39$\pm$20.53 \\
Cross-domain Single-positive CDVC & 80.36 & 7.53$\pm$5.44 & 70.08 & 9.14$\pm$7.51 & 60.61 & 10.07$\pm$6.98 \\
\bottomrule
\end{tabular}
\label{tab_ablation}
\end{table*}

\subsubsection{Results on the impact of the patch quadruplet generation strategy}
\label{subsec_abla_patch_gen}
We evaluated the effectiveness of our patch quadruplet generation strategy by comparing with three ablation variants: \textbf{1) Cropping-only:} only the random cropping pipeline was used; \textbf{2) Geometric-augmentation-only:} only the random geometric augmentation pipeline was used; \textbf{3) Cropping-and-geometric-augmentation:} the random cropping pipeline was applied followed by geometric augmentation.

Results of this ablation study are summarized in Table \ref{tab_ablation}-(a). As shown in this table, using only one pipeline led to consistent performance degradation. Specifically, compared to our proposed strategy, the cropping-only variant exhibited increases in MED of 2.04 mm, 3.89 mm, and 2.69 mm under the intra-dataset, cross-dataset intra-modality, and cross-dataset cross-modality settings, respectively. Similarly, the geometric-augmentation-only variant showed MED increases of 1.22 mm, 4.69 mm, and 1.38 mm under the three settings, respectively. These results indicate that a single pipeline is insufficient to capture the diverse spatial variations present in real data, thereby hindering effective discriminative voxel-wise embedding learning.

Furthermore, our proposed patch generation strategy consistently outperformed the cropping-and-geometric-augmentation variant across all one-shot evaluation settings, as shown in Table \ref{tab_ablation}-(a). In terms of MED, our strategy yielded lower errors, with reductions of 0.48 mm, 2.24 mm, and 1.76 mm relative to this variant under the intra-dataset, cross-dataset intra-modality, and cross-dataset cross-modality one-shot settings, respectively. For the ${\rm SDR}_{\rm 10 mm}$ metric, the improvements were 1.79\%, 9.47\%, and 8.88\%, respectively. These findings suggest that simultaneously applying large FOV shifts and severe geometric deformations, as in the combined pipeline, introduces overly destructive variations, which may prevent the network from establishing reliable voxel correspondences during contrastive learning.

\subsubsection{Results of sub-component ablations on the contrastive learning strategy}
To assess the effectiveness of the proposed contrastive learning strategy, we first examined the CPSC loss by removing both it and the DS decoder from the full model. Quantitative results are shown in Table \ref{tab_ablation}-(b). From this table, one can see that disabling semantic embedding learning led to consistent performance degradation, with MED increases of 3.84 mm, 6.25 mm, and 3.57 mm under the intra-dataset, cross-dataset intra-modality, and cross-dataset cross-modality one-shot settings, respectively. Meanwhile, reductions in ${\rm SDR}_{\rm 10 mm}$ of 21.87\%, 23.86\%, and 14.72\% were respectively observed. These results confirm that high-level semantic learning, effectively captured by the CPSC loss, is vital for precise one-shot landmark localization.

We next validated the efficacy of the CDVC loss, which employs a multi-positive selection strategy across multi-domain patch quadruplets. Two ablation variants were built, which used different pseudo-patches and positive selection strategies: \textbf{1) Intra-domain single-positive:} Only two intra-domain patches within each patch quadruplet were used to train the DD decoder and each anchor voxel was paired with a single positive drawn exclusively from the spatially corresponding area within the same domain; \textbf{2) Cross-domain single-positive:} Only two cross-domain patches within each patch quadruplet were used to train the DD decoder and each anchor voxel was matched with a single positive obtained from the spatially corresponding region across different domains. As presented in Table \ref{tab_ablation}-(b), it can be observed that our proposed multi-positive contrastive strategy consistently outperformed both ablation variants. In particular, compared to the intra-domain single-positive variant, the proposed strategy gained substantial improvements, especially under the cross-dataset intra-modality and cross-dataset cross-modality settings, where MED reductions of 12.75 mm and 14.74 mm, and ${\rm SDR}_{\rm 10 mm}$ improvements of 18.18\% and 37.66\% were respectively observed. When compared to the cross-domain single-positive variant, our strategy yielded MED reductions of 1.37 mm, 1.68 mm, and 0.42 mm, and ${\rm SDR}_{\rm 10 mm}$ improvements of 7.14\%, 7.95\%, and 4.11\% across the three evaluation settings, respectively. These results demonstrate that combining positive samples from both intra- and cross-domain patches provides more comprehensive supervision signals than single-positive strategies, which struggle to capture rich domain-invariant and discriminative features for robust generalizable one-shot landmark localization.

\section{Conclusion}
In this paper, we proposed an end-to-end trainable framework that learns UNIversal Voxel-wise Embedding via Randomized Synthesis, referred to as UNIVERS, for generalizable one-shot landmark localization in multi-modality volume scans of arbitrary FOVs. For model training, a domain-randomization-based data generation strategy was delicately designed to synthesize pseudo-image pairs and pseudo-patch quadruplets covering a wide range of appearances and spatial contexts. Built upon these pseudo-training data, a cross-domain voxel-wise discriminative contrastive learning strategy was then introduced to establish a domain-invariant embedding space that enables effective differentiation of anatomical structures. For accurate adrenal center localization at inference time, a coarse-to-fine one-shot localization framework was further developed by integrating a single frozen UNIVERS network. Extensive experiments across three challenging datasets, conducted under three one-shot settings with increasing difficulty, consistently demonstrated the superior performance of UNIVERS over the SOTA methods. These findings validate the strong generalization capacity of the proposed framework and position UNIVERS as a promising approach for one-shot landmark localization in real clinical applications.

 
 \bibliographystyle{IEEEtran} 
\bibliography{references}

\end{document}